\documentclass[preprint,12pt]{elsarticle}

\usepackage{amssymb}
\usepackage{xcolor}
\usepackage{overpic}
\usepackage{subfig}
\usepackage[margin=1in]{geometry}
\usepackage{amsmath}
\DeclareMathOperator{\NN}{NN}
\DeclareMathOperator{\ODESolve}{ODESolve}

\newcommand\norm[1]{\left\lVert#1\right\rVert}

\journal{XXX}

\begin{document}

\begin{frontmatter}

%% Title, authors and addresses

%% use the tnoteref command within \title for footnotes;
%% use the tnotetext command for theassociated footnote;
%% use the fnref command within \author or \address for footnotes;
%% use the fntext command for theassociated footnote;
%% use the corref command within \author for corresponding author footnotes;
%% use the cortext command for theassociated footnote;
%% use the ead command for the email address,
%% and the form \ead[url] for the home page:
%% \title{Title\tnoteref{label1}}
%% \tnotetext[label1]{}
%% \author{Name\corref{cor1}\fnref{label2}}
%% \ead{email address}
%% \ead[url]{home page}
%% \fntext[label2]{}
%% \cortext[cor1]{}
%% \affiliation{organization={},
%%             addressline={},
%%             city={},
%%             postcode={},
%%             state={},
%%             country={}}
%% \fntext[label3]{}

\title{Neural Lumped Parameter Differential Equations with Application in Friction-Stir Processing}

\author{James Koch\corref{cor1}, WoongJo Choi, Ethan King, David Garcia, Hrishikesh Das, Tianhao Wang, Ken Ross, Keerti Kappagantula}

\cortext[cor1]{Pacific Northwest National Laboratory, 902 Battelle Blvd., Richland, WA, 99354}

\begin{abstract}

Lumped parameter methods aim to simplify the evolution of spatially-extended or continuous physical systems to that of a ``lumped'' element representative of the physical scales of the modeled system. For systems where the definition of a lumped element or its associated physics may be unknown, modeling tasks may be restricted to full-fidelity simulations of the physics of a system. In this work, we consider data-driven modeling tasks with limited point-wise measurements of otherwise continuous systems. We build upon the notion of the \textit{Universal Differential Equation} (UDE) to construct data-driven models for reducing dynamics to that of a lumped parameter and inferring its properties. The flexibility of UDEs allow for composing various known physical priors suitable for application-specific modeling tasks, including lumped parameter methods. The motivating example for this work is the plunge and dwell stages for friction-stir welding; specifically, (i) mapping power input into the tool to a point-measurement of temperature and (ii) using this learned mapping for process control.

\end{abstract}

%%Graphical abstract
%\begin{graphicalabstract}
%\includegraphics{grabs}
%\end{graphicalabstract}

%%Research highlights
%\begin{highlights}
%\item Research highlight 1
%\item Research highlight 2
%\end{highlights}

%\begin{keyword}
%% keywords here, in the form: keyword \sep keyword

%% PACS codes here, in the form: \PACS code \sep code

%% MSC codes here, in the form: \MSC code \sep code
%% or \MSC[2008] code \sep code (2000 is the default)

%neural ordinary differential equations \sep lumped parameter \sep universal differential equation \sep friction stir welding \sep machine learning \sep control 

%\end{keyword}

\end{frontmatter}

%% \linenumbers

%% main text
\section{Introduction} \label{sec:intro}

Lumped-parameter methods aim to simplify the evolution of spatially-extended physical systems to that of a single ``lumped'' element. Such methods are common in  engineering and science where modeling spatially-extended physical systems may be unnecessary or intractable. Heat transfer is one such field that is amenable to such techniques: under certain assumptions and conditions, the flow of heat can be approximated by scaled temperature differences between bodies of uniform temperature. Thus, the modeling task is reduced to the evolution of a single unit under the action of energy fluxes across boundaries.

Many situations exist where lumped parameter methods may provide meaningful physical insights, including low-cost predictive capabilities, but the definition of the lumped element and its evolution may be unknown or ill-characterized. Typical concessions for proceeding with lumped-parameter modeling include neglecting property gradients within the modeled element and restricting input / output physics to occur along certain pathways (e.g. at a specific interface). 

In this work, we are motivated by \textit{Friction-Stir Processing} (FSP), whereby a non-consumable rotating tool is plunged into a workpiece, generating heat to process the surface of a workpiece without melting the material \cite{mishra2005friction}.  Final properties of the processed region are highly sensitive to the process conditions imposed such as temperature, force, plunge depth and the tool geometry. Typically, the required conditions are met through modifying the process parameters such as tool rotation rate, and tool traverse rate during FSP. Therefore knowledge of the mapping between process parameters and process conditions such as temperature behavior is critical for process control \cite{heidarzadeh2021friction}. However, modeling complex nonlinear systems like FSP is challenging. Both empirical and computational work has been done to understand the relationship between power and temperature, including computational fluid dynamics (CFD) approaches \cite{shrivastava2015physics,posada2012comparison,upadhyay2015thermal,mandal2008experimental,veljic2011numerical}. While many insights can be gained, detailed numerical models and empirical studies can incur significant costs, and high fidelity modeling may be unnecessary for applications. Here we consider a data-driven approach, and demonstrate it's performance using only a small set of experimental temperature measurements for FSP.

In practice, temperature during a FSP experiment is measured by a thermocouple embedded in the non-consumable rotating tool. We aim to learn the response of these measured temperatures in response to operator inputs using time series from past trial runs for processing a stainless steel alloy. The physics of this process is complex: fast physics of internal heat generation drives the flow of energy to cooler regions, which is dominated by the slower physics of gradient-dependent heat transfer. We construct a data-driven lumped-parameter model corresponding to thermocouple measurements of FSP tool temperature by leveraging assumed energy flow according to the First Law of Thermodynamics - that is, the conserved flow of energy - for a lumped element representative of the tool. With these inductive priors (lumped element obeying the first law of thermodynamics), we choose \textit{Universal Differential Equations} \cite{rackauckas2020universal} as the modeling paradigm upon which we build out models.

This paper is organized as follows: in Section \ref{sec:friction} we motivate the problem and desired modeling outcomes, including the FSP machinery, and instrumentation. In Section \ref{sec:ml}, we contextualize our work in the broader ML for Materials Science community. Our methods are described in Section \ref{sec:methods} with numerical experiments following in Section \ref{sec:exp}.

\section{Background} \label{sec:background}

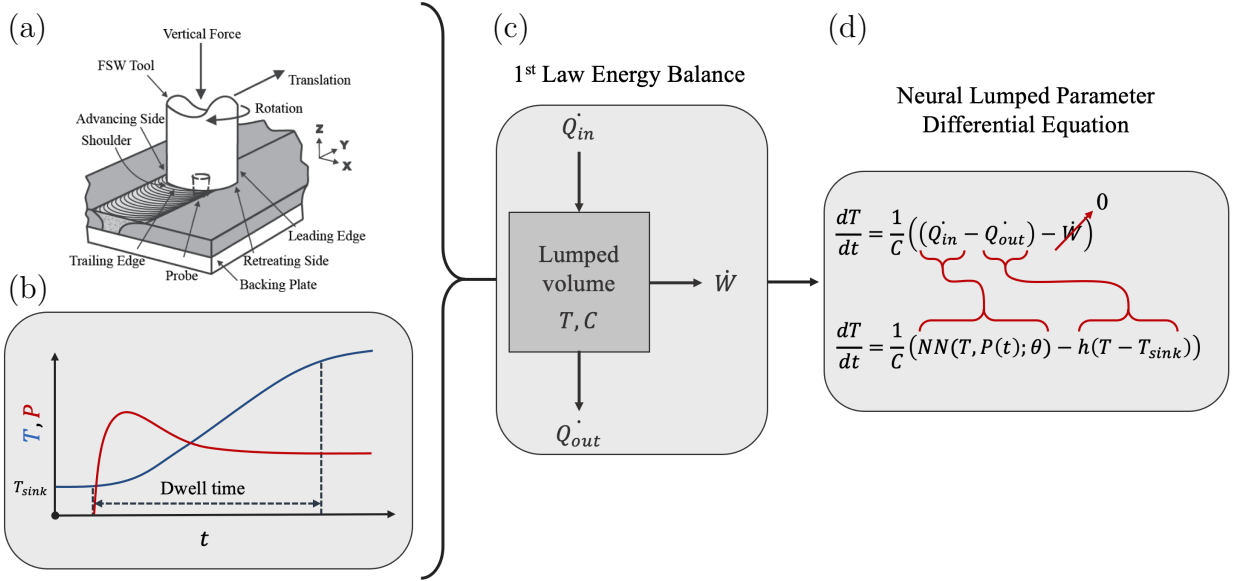
\begin{figure*}[t]
\centering
        \begin{overpic}[width=6.5in]{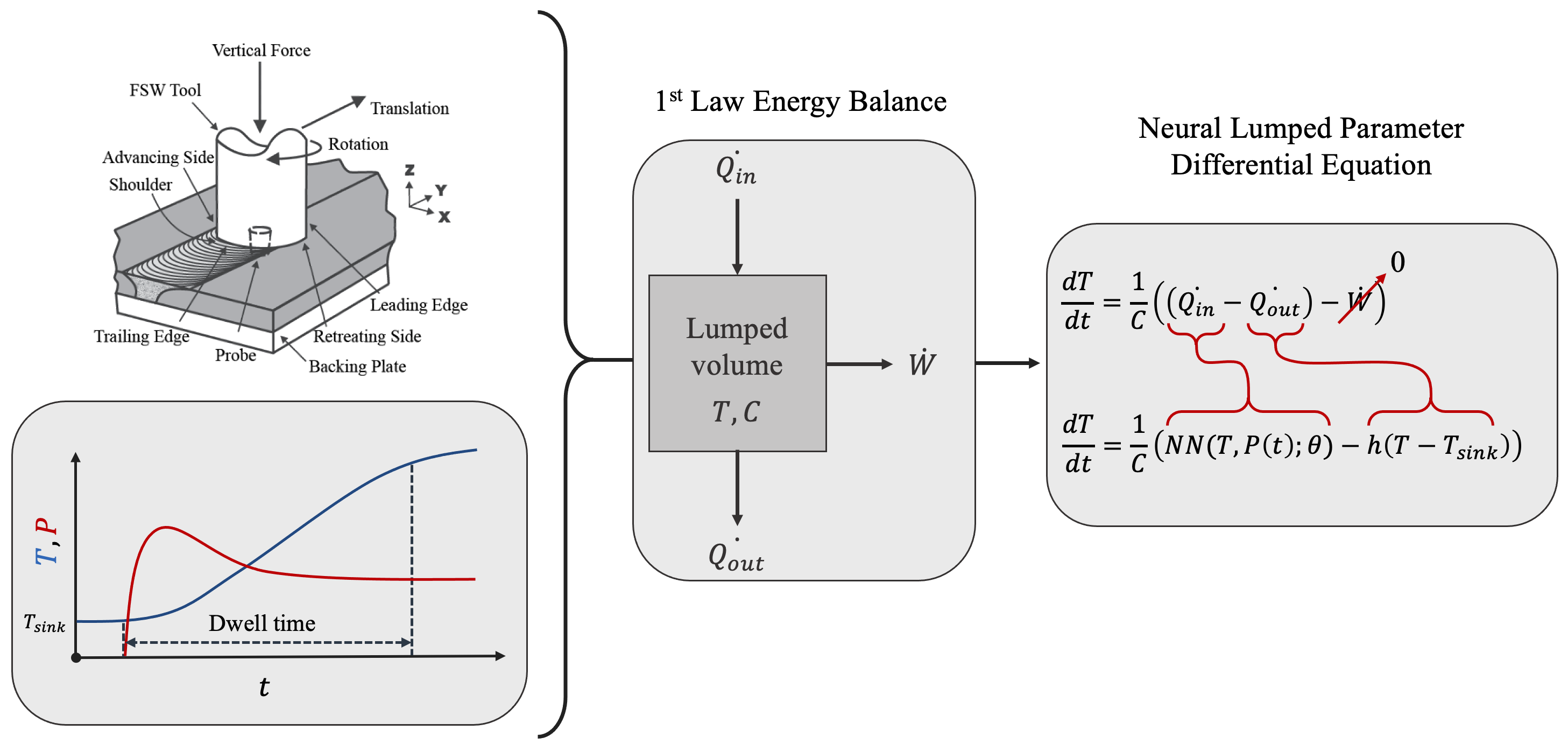}  
        \put(1,44){(a)}
        \put(1,23){(b)}
        \put(40,44){(c)}
        \put(67,44){(d)}
        \end{overpic}  
        \caption{\textit{Friction-stir processing} (FSP) utilizes internal heat generation from friction-based power transfer to plastically deform the  material for welding (a). In (b), notional time histories for temperature at the tool face and power input during the plunge and dwell stages of the welding are depicted. Once the temperature reaches a certain set point, the tool begins to travel across the workpiece. This work is concerned with modeling and control of the temperature profile during the plunge and dwelling stages. Our method is predicated upon the energy balance prescribed by the First Law of Thermodynamics (c). We construct an expected dynamical system, coined a \textit{Neural Lumped Parameter Differential Equation} (d), to impart this particular energy balance suitable for this problem.}
        \label{fig:overview}
\end{figure*}

\subsection{Friction Stir Welding} \label{sec:friction}

A relatively new joining processing, called \textit{Friction Stir Welding} (FSW), was invented at The Welding Institute (TWI) in the United Kingdom in 1991. \cite{thomas_friction_1995}  FSW utilizes the heat generated by the frictional surface contact between a rotating tool and workpiece that results in material softening and subsequent severe plastic deformation along with material mixing to bond materials together. The entire process takes place in the solid-state, where the processing temperature remains below the melting temperature. Due to the absent of excessive heat and melting, FSW processes are known as energy efficient methods in comparison to arc welding, where materials are melted for bonding. In addition, FSW welds do not experience problems with re-solidification, such as cracking, porosity, and embrittlement. For theses reasons, friction stir welding has been an attractive joining process in many industries. FSP is a derivative approach to FSW where there is only one workpiece that is modified through the plunging and traverse of the non-consumable tool.      

The FSP process involves four stages: plunging, dwelling, traversing, and extracting. The initial plunge stage refers to the period where the rotating tool is in contact with the workpiece with downward directional force, referred to as the plunge force. The dwelling stage refers to the period where rotational friction continues, but there is no additional downward directional motion. The material plastically deforms and heats up during the plunge and dwell stages. Then during the traverse stage the rotating tool moves along the a defined path. The plasticized material is mixed and extruded past the rotating and traversing tool.  Lastly, at the end of the traverse line, the rotating tool is extracted, leaving the exit hole. 

The plunge and dwelling stages in FSP process are extremely important, since most of the initial thermo-mechanical conditions are generated and the workpeice undergoes significant material transformation due to the rapid temperature increase and forging pressure \cite{mandal2008experimental}. During these stages, materials undergo significant transformation from a solid state to a plastically deformed condition in a short period of time. Due to the rapid condition change, most of the tool wear occurs during these stages \cite{thomas_friction_1995,lienert2003friction,mandal2006thermomechanical}. Unsuccessful temperature control at the end of the dwell stage often leads to overshoot and undershoot of the temperature during the weld traverse. This may cause stability issues even under well tuned temperature control algorithms \cite{ross2012investigation,taysom2017comparison}. A thorough understanding of the temperature profile prior to the traverse stage is critical to preserving tool life and securing optimal process conditions during the traverse stage.  

\subsection{ML for Temperature Prediction} \label{sec:ml}

Much work has been done to construct high fidelity models of FSW and FSP processes, often leveraging computational fluid dynamics approaches ~\cite{Bussetta2014,Ulysse2002,Colegrove2006,Bastier2006,Jain2016,Chiumenti2013}. In general these approaches require significant computational resources that can be prohibitive for their use in process design and control. To lessen this burden, data driven and machine learning methods have been utilized to learn relationships between process parameters and weld properties from limited numerical or computational experiments that can then be used in process design. For example, impacts of tool geometry \cite{Shojaeefard2015} and chemical composition \cite{Li2020}. However, less work has been done to construct simplified models of FSW dynamics for the purposes of process control. At the level of process control, the detail of high fidelity models may be unnecessary, but it is challenging to construct simplified dynamics that can accurately predict measured outputs such as temperature. In a related solid phase process called shear assisted process extrusion (ShAPE), Wight et al. developed DeepTemp, a recurrent neural network that can accurately predict measured temperature during processing given the operator inputs \cite{Wight2023}. A simplified thermodynamic model augmented with a neural network has also been shown to produce accurate predictions of process temperature dynamics for ShAPE \cite{King2023}. Addition of physics can allow for greater interpretability of learned dynamics and allow for greater confidence in their use for process control.

\section{Methodology} \label{sec:methods}

\subsection{Lumped-parameter methods}

Lumped parameter models reduce complex systems with infinite degrees of freedom (e.g. spatially-varying fields) to a single degree of freedom that is representative of the whole unit, thereby reducing space-time evolution of a governing Partial Differential Equation (PDE) to the time evolution of a system of Ordinary Differential Equations (ODE). The prototypical example is that of a lumped thermal unit; for example, a room in a building whose state can be represented by a single temperature that evolves according to its capacitance (i.e. thermal capacitance associated with the volume of the room) and resistive coupling to other rooms. In heat transfer problems, one expects temperatures to behave as a first-order system; i.e., temperatures typically asymptotically approach a steady state value where the energy input to the system is eventually matched by the sum of the dissipative physics of the problem. 

In FSP, the workpiece is heated through power input via a spinning tool. On the time scales relevant to FSP, heat dissipation is primarily through conduction away from the stir region. With the notional dominant balance physics in mind, one can construct a framework around these energy pathways consistent with the time-dependent First Law of Thermodynamics:

\begin{equation} \label{eq:first_law}
    \frac{dE}{dt} = \dot{Q} - \dot{W},
\end{equation}
where $E$ is the energy contained in the system, $\dot{Q}$ is the rate of heat input to the system, and $\dot{W}$ is the rate of work done by the system on its surroundings. To simplify the problem further, we define the boundaries of the assumed lumped-volume to be placed such that there is no work exchange between the modeled system and its surroundings. Thus, the work term can be neglected and the energy balance for the system becomes the competition between energy gain and loss:
\begin{equation} \label{eq:heat_balance}
    \dot{Q} = \dot{Q}_{in} - \dot{Q}_{out},
\end{equation}
Substituting Eq. \ref{eq:heat_balance} into Eq. \ref{eq:first_law} and recasting energy dynamics and thermal dynamics, we obtain:
\begin{equation} \label{eq:lumped_pathway}
    \frac{dT}{dt} = \frac{1}{C}\left( \dot{Q}_{in} - \dot{Q}_{out} \right),
\end{equation}
with $C$ representing the thermal capacitance of the lumped volume. Thus, for a well-defined lumped volume, the thermal dynamics are reduced to equilibrating heating and cooling physics. 

\subsection{Neural Ordinary Differential Equations}

\textit{Neural Ordinary Differential Equations} (NODEs \cite{chen2018neural}) are a machine learning modeling paradigm where the transition dynamics of a system are approximated by a tunable function approximator; e.g. a trainable neural network. NODEs differ from other ML time series modeling methods (e.g. RNNs, LSTMs, etc.) in that (i) the treatment of time is continuous as opposed to discrete timesteps, (ii) the dynamics can be made interpretable by the inclusion of various physical priors, and (iii) NODEs can leverage standard ODE integration techniques (e.g. adaptive step Runge-Kutta methods or integrators for stiff dynamics).

Neural ODEs evolve a system's state $x$ through an independent variable (typically time $t$) through the solution to the initial value problem (IVP):
\begin{equation} \label{eq:ivp}
    x_{t_{end}} = x_{t_0} + \int_{t_0}^{t_{end}} f\left(x,t;\theta\right) dt ,
\end{equation}
where $f(\cdot)$ is the learnable Right-Hand-Side (RHS, or the transition dynamics) of the ODE parameterized by the parameter vector $\theta$. We denote the solution to this IVP as:
\begin{equation}
    x_{t_{end}} = \ODESolve \left( f\left(x,t;\theta\right), x_0, t_0, t_{end} \right),
\end{equation}
that is, the solution to the ODE from the initial condition $x_0$ for the temporal span of $t_0$ to $t_{end}$. Note that as written, this is a non-autonomous differential equation because it depends on the independent variable $t$. In general, the model can be autonomous or non-autonomous depending on the specific application. The IVP solution can be made differentiable one of many ways: (i) backpropagation through the elementary operations of the solver (reverse-mode auto-differentiation), (ii) through the adjoint sensitivity method, or (iii) through forward sensitivity analysis (forward-mode auto-differentiation). 

The transition dynamics of $f$ can be made of any differentiable mapping allowing for the inclusion of known physical priors or universal function approximators, such as feed-forward neural networks. In this manner, Neural ODEs can exist as `black-box' models, where a learnable dynamical system is expressed as:
\begin{equation}
    \frac{dx}{dt} = \NN(x,t;\theta),
\end{equation}
with $\NN(\cdot)$ representing the tunable Neural Network parameterized by $\theta$. The inclusion of various physical priors allows for gray- and white-box modeling tasks, such as surrogate and/or closure modeling and parameter tuning problems. This class of domain-informed neural ODEs is called \textit{Universal Differential Equations} \cite{rackauckas2020universal}.

\subsection{Neural Lumped-Parameter Differential Equations}

In this work, we focus on modeling the input and output energy balance present in the friction-stir welding process through fitting a differential equation of the form of Eq. \ref{eq:lumped_pathway} to the available experimental data. We propose a sub-class of neural ODEs called \textit{Neural Lumped-Parameter Differential Equations} constructed to exploit knowledge of input and output energy pathways and capacitive first-order physics. At this level of abstraction, we expect to fit models (for temperature of a lumped volume) of the form:

\begin{equation} \label{eq:lumped_node}
    \frac{dT}{dt} = \frac{1}{C} \left( f\left( T,t;\theta \right) - g\left( T,t;\phi \right) \right),
\end{equation}
with $f(\cdot)$ and $g(\cdot )$ representing (potentially time-dependent) learnable input and output heat transfer pathways, respectively.

Imparting such domain knowledge acts to (i) regularize the regression to promote training stability and generalizablility of the model and (ii) increase the interpretability of the model for downstream tasks, such as control and/or analysis. In the context of FSP, we can further inform the design of the input and output energy pathway functions from domain knowledge. First, we assume that the loss of energy from the domain is linear with temperature difference between that of the volume and its assumed-constant surroundings; i.e.:
\begin{equation}
    g\left( T \right) = h\left(T - T_{sink} \right) ,
\end{equation}
where $h$ is the heat transfer coefficient and $T_{sink}$ is the pseudo-infinite sink temperature. Both $h$ and $T_{sink}$ can be user-specified or learned. With this specified energy loss term, we aim to ``close'' the model by finding an appropriate approximation of the energy input by tuning a feed-forward neural network $f = \NN\left(T,P\left(t\right);\theta\right): \mathbb{R}^2 \rightarrow \mathbb{R}^1$, where $P(t)$ is the recorded input power over time. Thus, the final model form that we aim to fit is:
\begin{equation}
    \frac{dT}{dt} = \frac{1}{C} \left( \NN\left(T,P\left(t\right);\theta\right) - h\left(T - T_{sink} \right) \right) = M(T,P(t);\Theta),
\end{equation}
where $\Theta$ denotes the inclusive set of learnable parameters in the model. Note that these modeling choices were made to simplify the modeling procedure by reducing the number of tunable parameters. Should more complex physics be present, the energy input and output functional forms can be recast for increased flexibility (e.g. deep neural networks). 

Known, for example, is that the time constants for the thermocouple response can vary from stage-to-stage. In the initial plunge and dwell stages, the time constant may be larger than that of traversal: the input and output energy balance is not in equilibrium and leads to a first-order rise in temperature as measured at the embedded thermocouple. Once a quasi-steady state condition has been reached and tool traversal has begun, the input and output energy balance is altered due to the movement of the tool. We explicitly restrict our study to the plunge and dwelling stages of the presented experiments. To generalize to other stages would require (i) further parameterization of the model or (ii) implementing piece-wise continuous models trained on separate processing stages.

\section{Friction-Stir Processing Experiments} \label{sec:hardware_exp}

The data used to train the model was sampled from 34 inch bead on plate FSP experiments of 316L stainless steel plate. Specifically, time series measurements of the mechanical input power and temperature during the plunge and dwelling stages were utilized to train the model. The mechanical power input is estimated by the product of spindle speed and torque from the spindle motor. An enconder measures spindle speed of the motor and torque is calibrated from the motor current. The 316L plates were commercially available hot rolled, solution annealed at 1080 $^{\circ}$C and water cooled. The chemical composition of the as-received plates provided was: C: 0.029 wt. \%, Cr: 18.180 wt. \%, Ni: 8.029 wt.\%, Mn: 1.871 wt. \%, Si: 0.281 wt. \%, S: 0.001 wt. \%. The plates were machined to have a pilot hole to minimize excessive flash at the start of the weld. Mazak MegaStir tools made of polycrystalline cubic boron nitride (PCBN) with 30 wt \% W-Re tool were used.  A embedded K- type thermocouple is place at the back of PCBN tool and the temperature data is transferred wirelessly through a temperature transmitter to data acquisition at 10 Hz rate. 

\section{Temperature Modeling} \label{sec:exp}
\begin{figure}[t]
        \centering
        \begin{overpic}[width=5in]{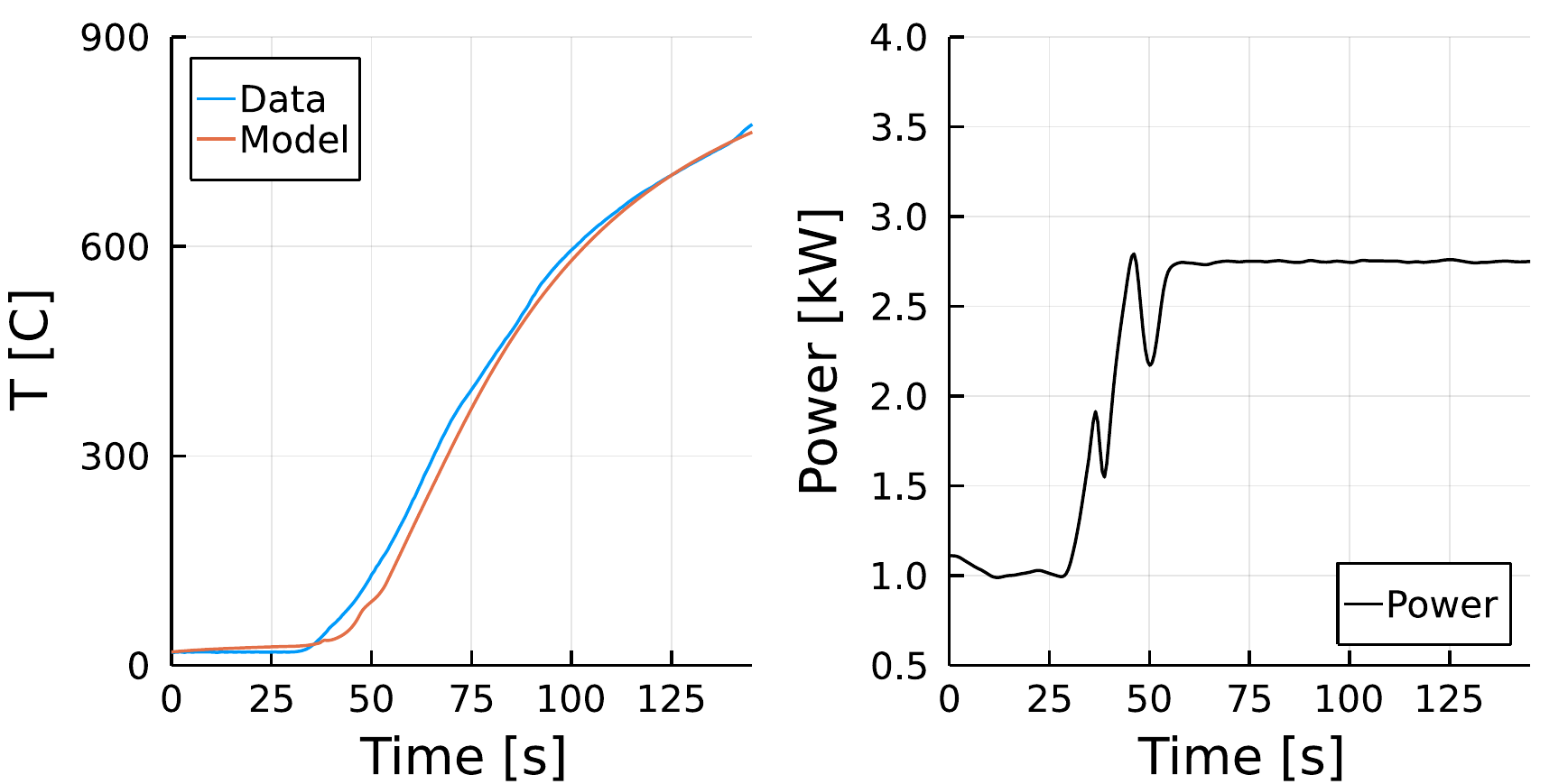}  
        \put(1,52){(a)}
        \put(50,52){(b)}
        \end{overpic}  
        \caption{In (a), shown is an example thermocouple-recorded temperature profile during the dwelling phase of a friction-stir welding run. The corresponding Neural Lumped Parameter Differential Equation model solution is shown in red. In (b), the experimental control input that produced the time series in (a) is shown. This control input is used during training to construct a surrogate mapping from the control signal and current temperature to internal heat generation.}
        \label{fig:results}
\end{figure}

\begin{figure}[t]
        \centering
        \begin{overpic}[width=5in]{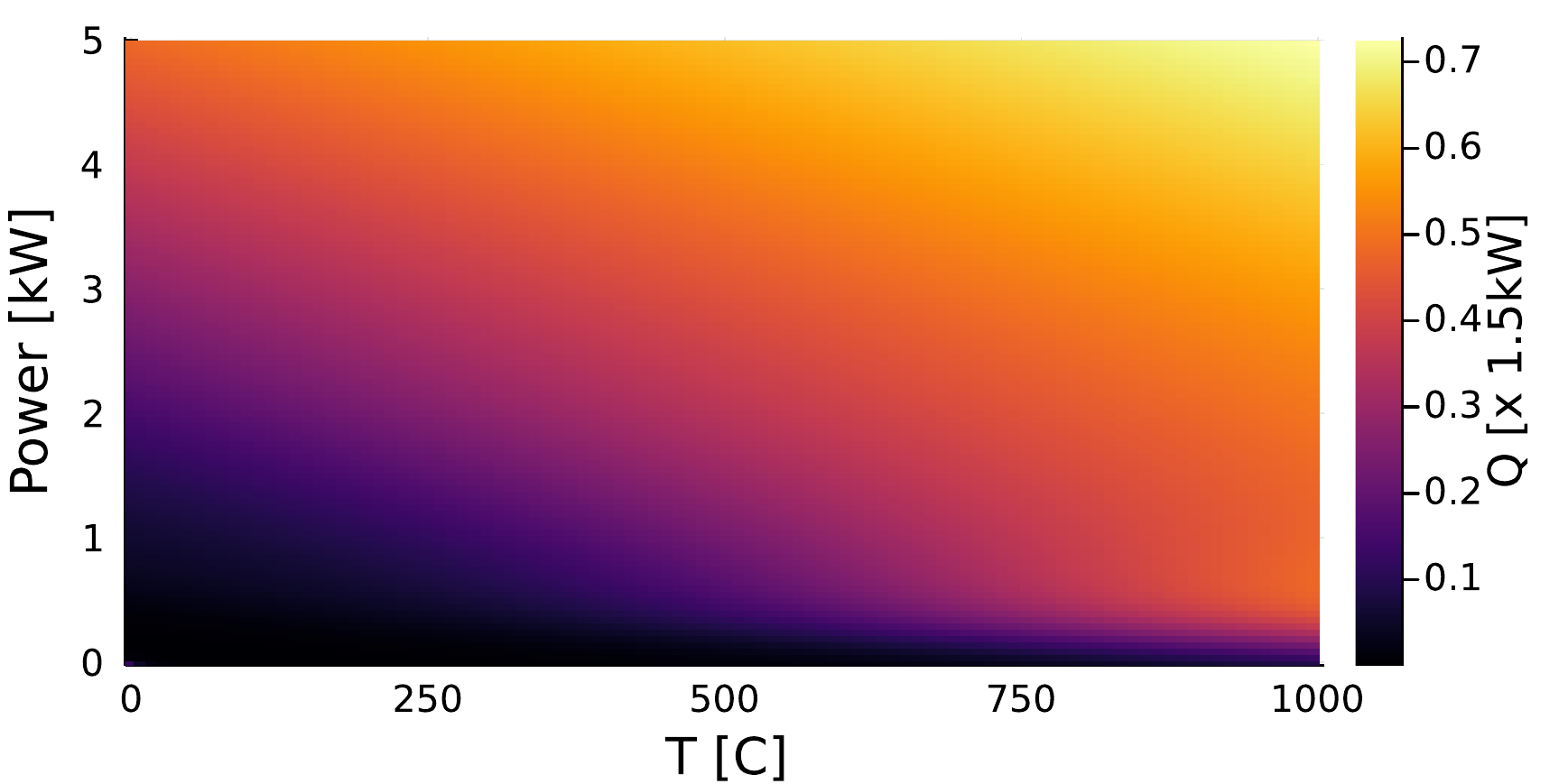}  
        \end{overpic}  
        \caption{Upon successful training of the neural lumped parameter model (Eq. \ref{eq:lumped_node}), the learned surrogate for internal heat generation can be queried to give an estimate of the mapping between input power, temperature, and resultant heat generation.}
        \label{fig:heatmap}
\end{figure}

\begin{figure*}[t]
        \centering
        \begin{overpic}[width=5in]{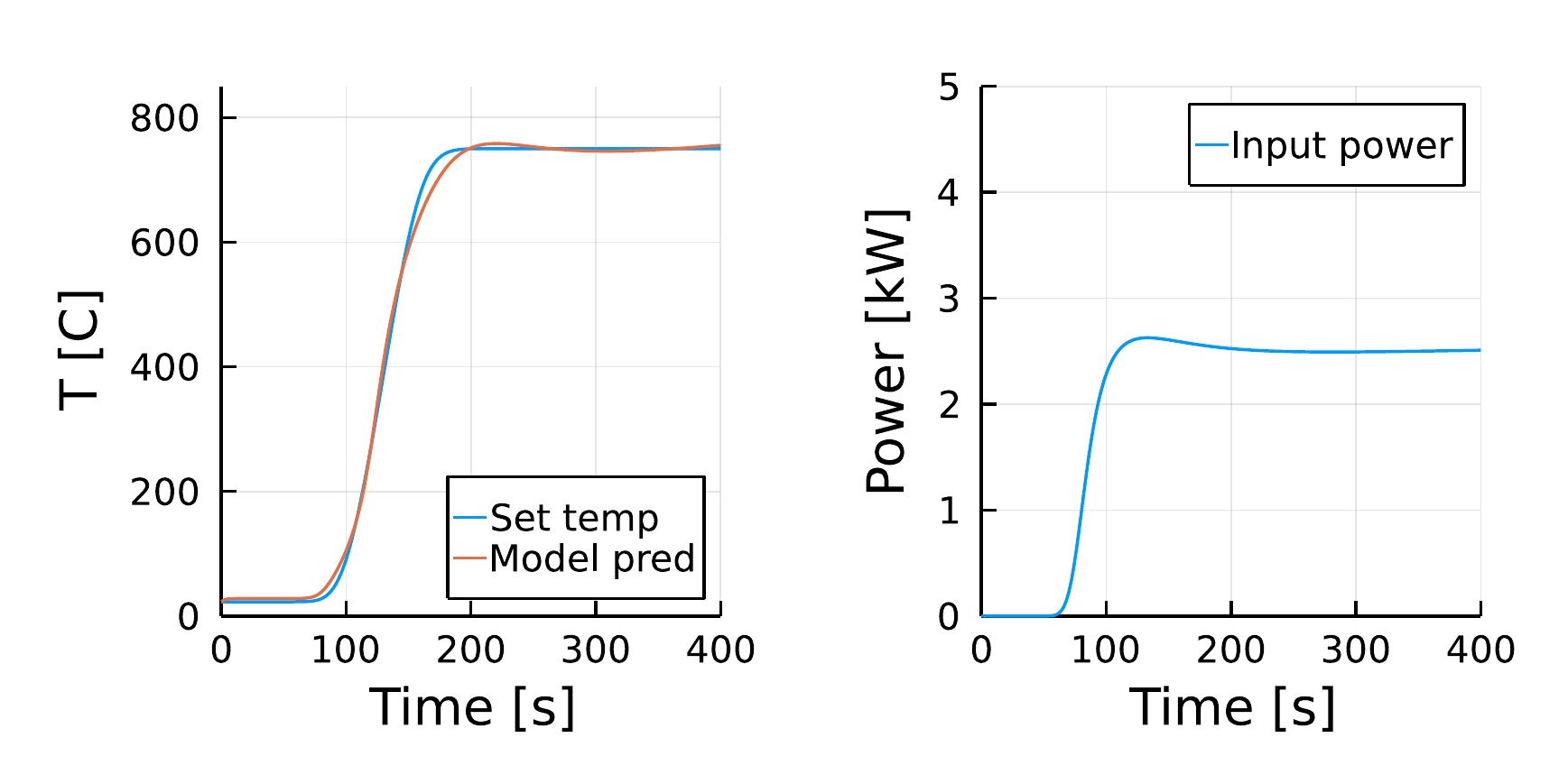} 
        \put(3,42){(a)}
        \put(53,42){(b)}
        \end{overpic}  
        \caption{The system identification task shown in Fig. \ref{fig:results} allows for investigation of different model-based control strategies, e.g. Model Predictive Control (MPC). In (a), the goal is to reach and hold a set temperature of $700$ Celsius. In (b), an optimization-obtained control input (achieved through power input at the friction-stir welding tool) is shown for a system with a capped maximum power input of 4 kW. }
        \label{fig:control}
\end{figure*}

\subsection{Model Setup}

We seek to minimize the mean-squared error between model time series data $\hat{y}_{i,j}$ and experimentally-obtained temperature time series data $y_{i,j}$ over a collection of experimental runs; i.e. minimize the loss function:
\begin{equation} \label{eq:loss}
    \mathcal{L} = \frac{1}{N_i} \sum_{i}^{N_i} \sum_{j}^{N_j}  \left(\hat{y}_{i,j} - y_{i,j}\right)^2,
\end{equation}
where $N_i$ is the number of unique experimental runs in the training data set and $N_j$ is the number of equally-spaced data points for each experimental run. The model data $\hat{y}_{i,j}$ is defined by the forward-pass of the model over each of the experimental runs:
\begin{equation}
    \hat{y}_{i,j} = \ODESolve( M(T,P(t);\Theta),T_0,t_0,t_{end})|_{t=j \cdot \Delta t},
\end{equation}
that is, the solution to the $i$-th IVP recorded at the co-location points $j$ specified by the fixed timestep $\Delta t$. 

For modeling temperature specifically, we use a scaled feed-forward neural network $Q_0 \cdot \NN(T/T_0,P/P_0): \mathbb{R}^2 \rightarrow (0,Q_0) $ mapping input temperature and power to internal heat generation with one hidden layer of 10 nodes (`swish'-activated) and a sigmoidal output layer. Note that although we do not non-dimensionalize the model, we do scale the associated physical quantities such that they are all approximately the same order of magnitude. Likewise, the neural network is bounded to the interval $(0,Q_0)$ as a heuristic to enforce physicality for the learned heat generation surrogate. Similarly, we select $T_{sink}$ to be fixed at room temperature, $23$ Celsius, and set the heat transfer coefficient to $h=1$. Thus, in total, the parameter vector $\Theta$ is comprised of the neural network weights and biases $\theta$ and the capacitance of the lumped volume $C$.

We perform the regression in the Julia computing ecosystem, especially leveraging the packages DifferentialEquations.jl \cite{rackauckas2017differentialequations} and DiffEqFlux.jl \cite{rackauckas2020universal}. The ODE solver is an adaptive fourth-order Runge-Kutta integrator. For each optimization task, we use Adam for 200 epochs with a learning rate of 0.001 followed by the BFGS optimizer until converged (marked by small relative change in the loss between training epochs). 

\subsection{Low-data Training and Validation}

Seven FSP experiments were recorded (i.e. having full temperature time histories) with consistent material selections and similar dwell times. Given this limited amount of data, and with a goal of training an interpretable and generalizable model, we employ a data augmentation strategy and a low-data model validation strategy.

A risk in modeling a low number of unique time histories is that the model can `memorize' the shape of the traces and / or their locations in time. In addition to the lumped parameter prior built into the model by construction, we augment the data with additional copies of the training data that have been randomly shifted in time. In this manner, the model is encouraged to learn the appropriate input and output energy physics during the tool plunge and dwell stages.

For model reporting, we performed a 10-fold shuffle-split training strategy. In each iteration, four training trajectories were sampled from the set of seven and the remaining three were withheld as a test set. 
\subsection{Results}

Table \ref{tab:summary} summarizes the results of the model training and reporting tasks. An example reproduction of time series data for a particular run is shown in Fig. \ref{fig:results} with weights corresponding to Trial \#1. A comprehensive set of results showing the qualitative differences between the `best' model and the `worst' model (as judged by the magnitude of the test loss) is listed in the Appendix.

\begin{table}[h!]
\centering
\caption{Summary of training task. \textit{Train} and \textit{Test} magnitudes correspond to the loss function (Eq. \ref{eq:loss}). $C$ is the identified capacitance for the assumed lumped volume for each trial.} 
\label{tab:summary}
\begin{tabular}{lcccccccccc} 
Trial No. & 1 & 2 & 3 & 4 & 5 & 6 & 7 & 8 & 9 & 10 \\ \hline
Train &  820. & 351. & 227. & 704. & 286. & \textbf{115.} & 199. & 321. & 655. & 160. \\
Test & \textbf{974} & 3110 & 4730 & 1020 & 2000 & 1520& 1320 & 1260 & 1160 & 1260\\
$C$ &3.97 & 3.93 & 3.44 & 2.93 & 3.94 & 17.4 & 18.1 & 3.75 & 3.87 & 6.10 \\ 
\end{tabular}
\end{table}

In all training scenarios, the model was able to reproduce the characteristic first-order rise of temperature seen in the experiments. Furthermore, as evidenced by the results in Table \ref{tab:summary}, the learned capacitance values are all of the same order, with many converging to a similar value. Figure \ref{fig:heatmap} depicts the response surface of the learned surrogate model (trained neural network) mapping temperature (horizontal axis) and input power (vertical axis) to internal heat generation (color in the heatmap).

\section{Control}
A trained model of the true FSP temperature response to power input is immediately useful for informing operation of the tool during processing. The learned model parameters can be held fixed and control inputs can then be learned to achieve target outcomes. We demonstrate these techniques by constructing a power profile for the initial plunge stage such that temperature rises smoothly to a target set point. The trained lumped-parameter model is only assumed to be valid in the plunge and dwell stages. A smooth transition between an imposed power profile and temperature feedback control (e.g. PID) can occur for subsequent stages. Construction of a data-driven control in this manner has the potential to reduce overshoot and settling times to target set temperatures.

\subsection{Model Setup}
Beginning from the model of Eq. \ref{eq:lumped_node}, we replace the power signal $P(t)$ with a second feed-forward neural network mapping time to power; i.e. $P_0 \cdot NN(t;\phi): \mathbb{R}^1 \rightarrow (0,P_0)$, with $P_0$ representing the maximum power. The neural network has two hidden layers with 5 nodes, each `swish' activated, with a sigmoidal output layer. With all other model parameters fixed, we train this second neural network based on the loss function:
\begin{equation} \label{eq:control}
    \mathcal{L} = \norm{T_{set} - T}_2^2,
\end{equation}
that is, minimizing the difference between the model trajectory and the reference trajectory. The training procedure uses the Adam optimizer for 200 epochs and then the BFGS optimizer for fine tuning until converged.

\subsection{Results}
Figure \ref{fig:control} depicts an optimized solution to Eq. \ref{eq:control} for a set temperature of 700 celsius and maximum power input of $4kW$. After an initial steep rise in power over an approximately 60-second period, the power level steadies to a constant power required for the steady-state input and output energy balance.

\section{Discussion and Conclusions}

This paper presented methodology for modeling characteristic first-order-rise temperature time series from FSP experiments performed on 316L steels. Consistent with engineering heat transfer modeling techniques, we extended the notion of the lumped-parameter method to include tunable physics to be learned from data directly. The regression task was built within the \textit{Universal Differential Equation} paradigm, where one fits a dynamical system to data. In this work, we construct the model dynamical system to contain the same input and output physics present in the friction-stir welding experiment, namely internal heat generation and heat transfer away from the weld region. The lumped-capacitance abstraction is sufficient to capture essential physics for reproducing behavior seen in time series data (Fig. \ref{fig:results}). Because the model is constructed with constituent energy pathways, the model is interpretable in that one can use the learned closure (mapping from tool force and temperature to heat input) for inference tasks; i.e. predictive control as shown in Fig. \ref{fig:control}. 

As presented, the learned models do not generalize beyond this specific experimental setup, including material choices. A natural extension of our work is to include parametric dependence of the experiments during training of the models. For example, including material density, specific heat capacity, and heat transfer coefficients can help craft a higher-fidelity response surface (Fig. \ref{fig:heatmap} ) tailored to a specific experimental campaign. These higher-dimensional parameterizations can help dramatically reduce the need for costly high-resolution grid-search strategies in experiments.

\section*{Acknowledgements}

The research described in this paper is part of the Materials Characterization, Prediction, and Control agile investment at Pacific Northwest National Laboratory. It was conducted under the Laboratory Directed Research and Development Program at PNNL, a multiprogram national laboratory operated by Battelle for the U.S. Department of Energy under contract DE-AC05-76RL01830.

\section*{Author Contributions}
The authors confirm contribution to this manuscript as follows: study conception and design: J. Koch, W. Choi, E. King; analysis: J. Koch; data creation: D. Garcia, H. Das, T. Wang, K. Ross, K. Kappagantula; data curation: W. Choi; draft preparation: J. Koch. All authors reviewed the manuscript and approved its final version.

\bibliographystyle{elsarticle-num} 
\bibliography{main}

%% The Appendices part is started with the command \appendix;
%% appendix sections are then done as normal sections
\newpage
\appendix

\section{Complete Results}
\label{app:Results}

\begin{figure*}[ht!]
        \subfloat[]{%
            \includegraphics[width=.48\linewidth]{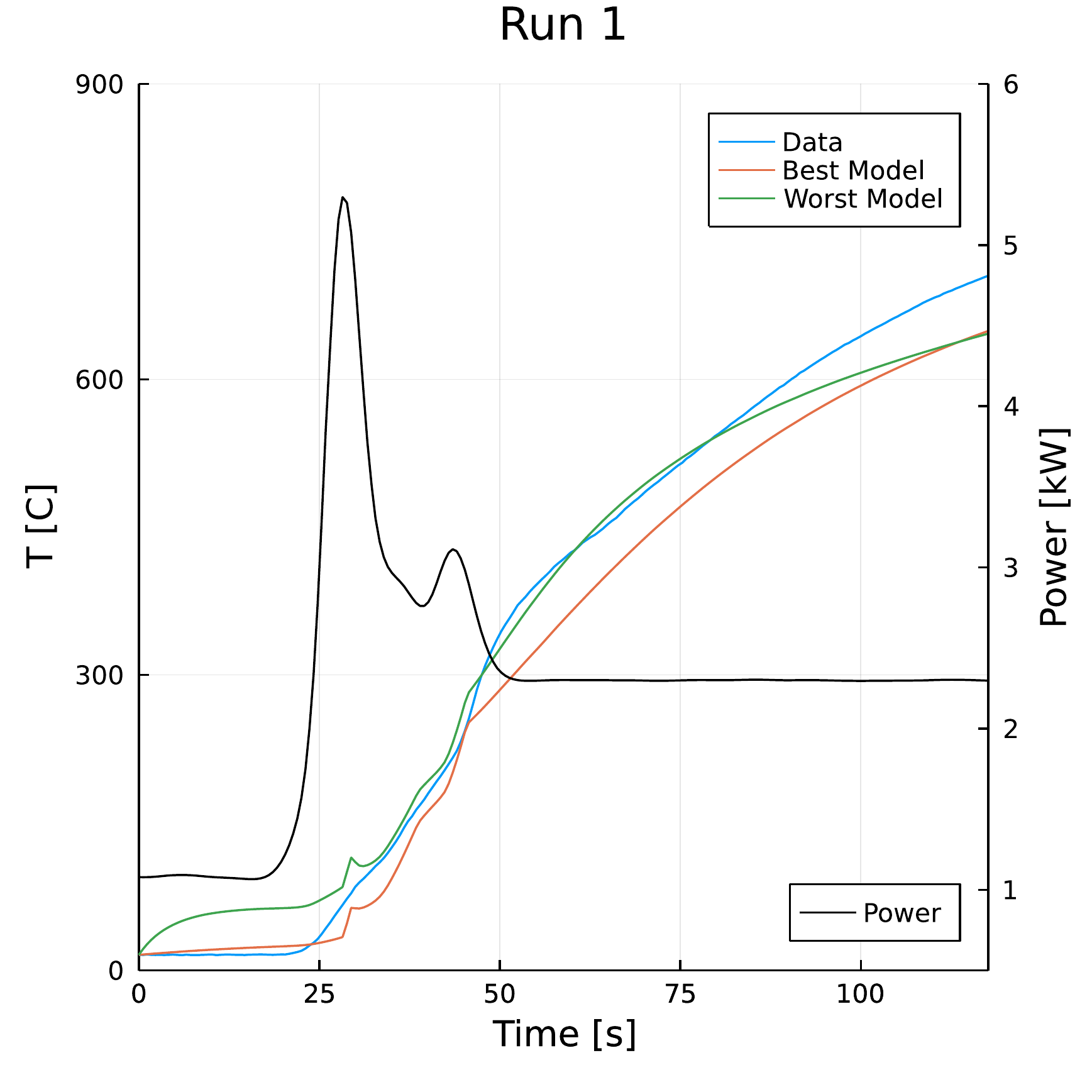}%
            \label{subfig:1a}%
        }\hfill
        \subfloat[]{%
            \includegraphics[width=.48\linewidth]{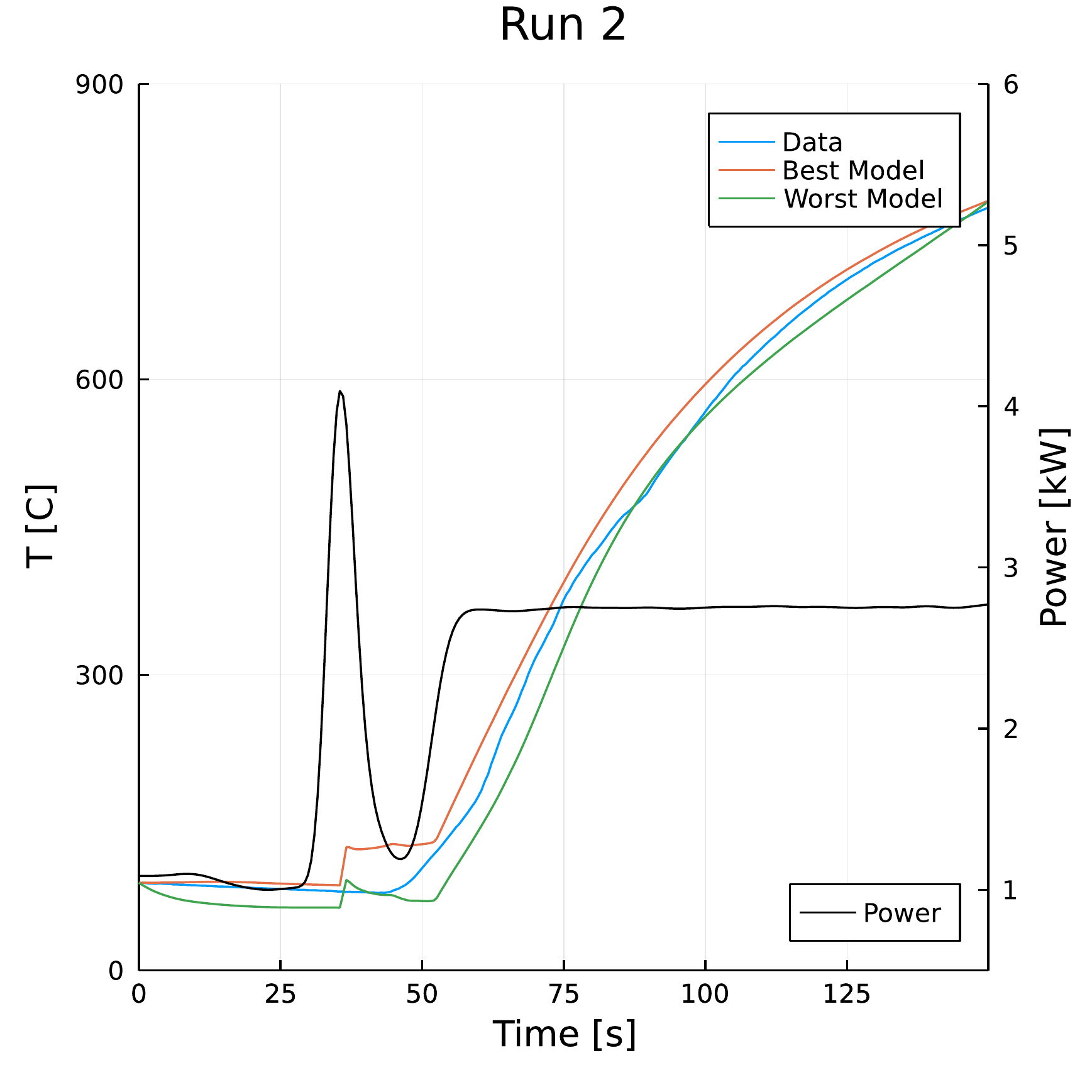}%
            \label{subfig:1b}%
        }\\
        \subfloat[]{%
            \includegraphics[width=.48\linewidth]{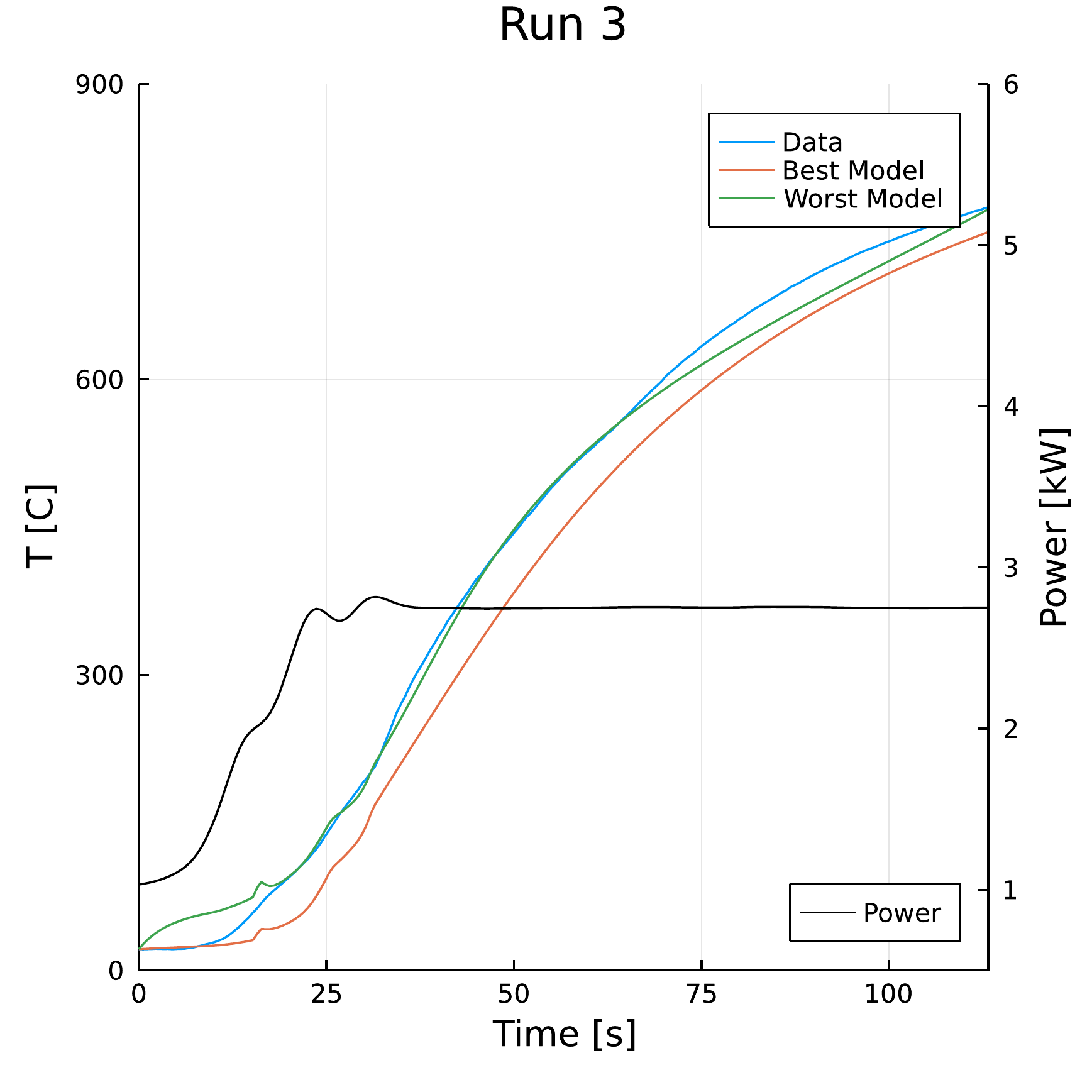}%
            \label{subfig:1c}%
        }\hfill
        \subfloat[]{%
            \includegraphics[width=.48\linewidth]{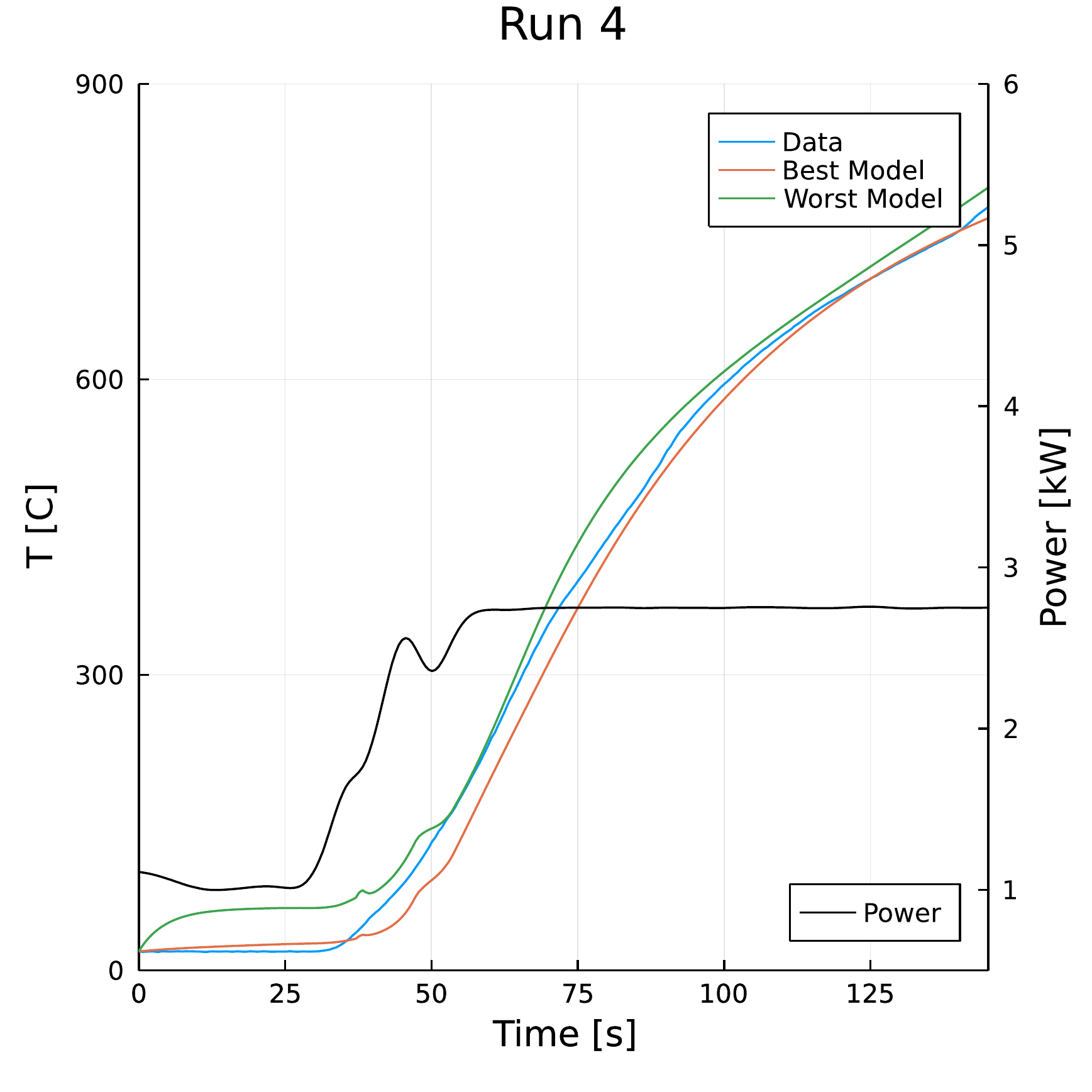}%
            \label{subfig:1d}%
        }
        \caption{Full results for experiments \#1 through \#4.}
        \label{fig:results_summary_1}
    \end{figure*}

\begin{figure*}[ht!]
        \subfloat[]{%
            \includegraphics[width=.48\linewidth]{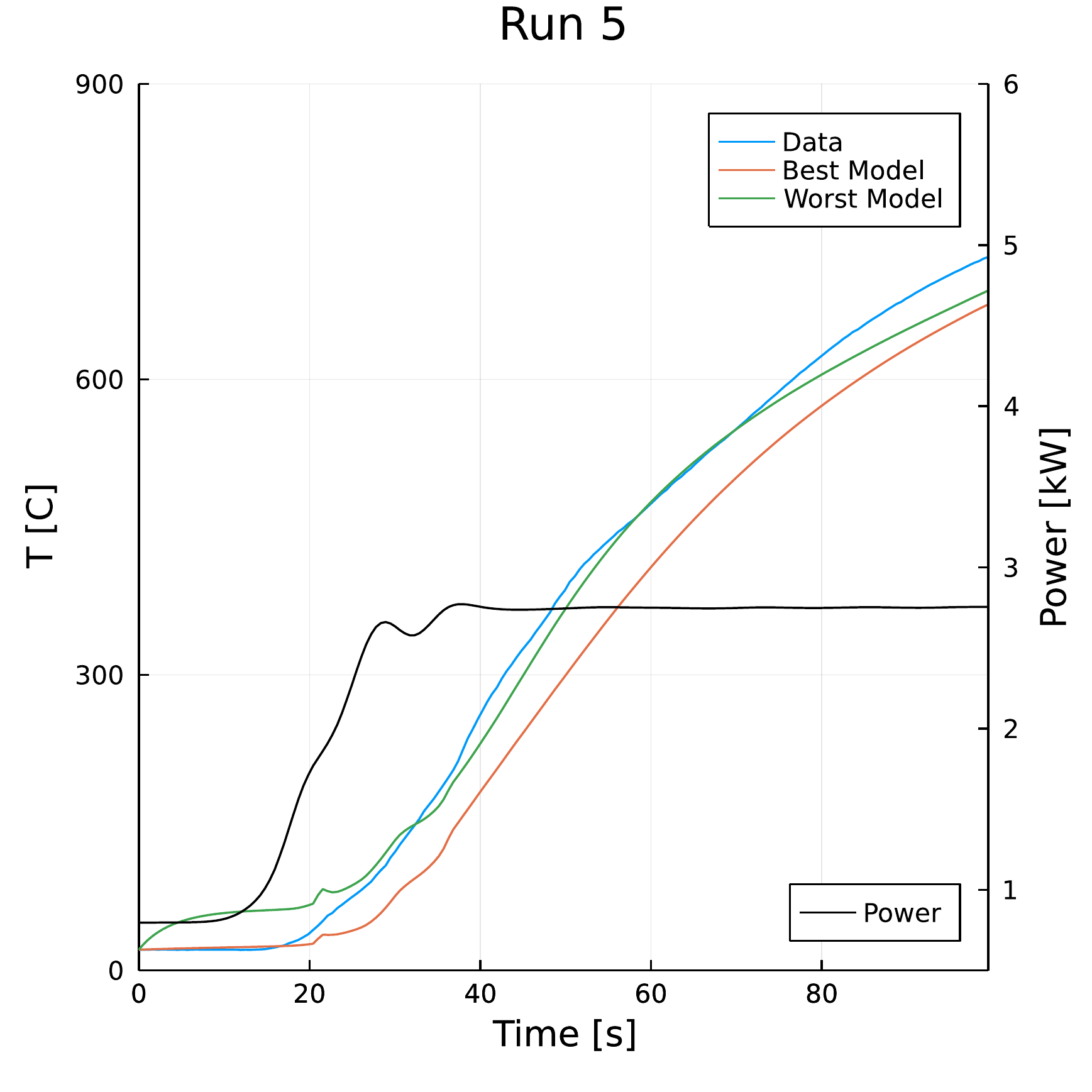}%
            \label{subfig:2a}%
        }\hfill
        \subfloat[]{%
            \includegraphics[width=.48\linewidth]{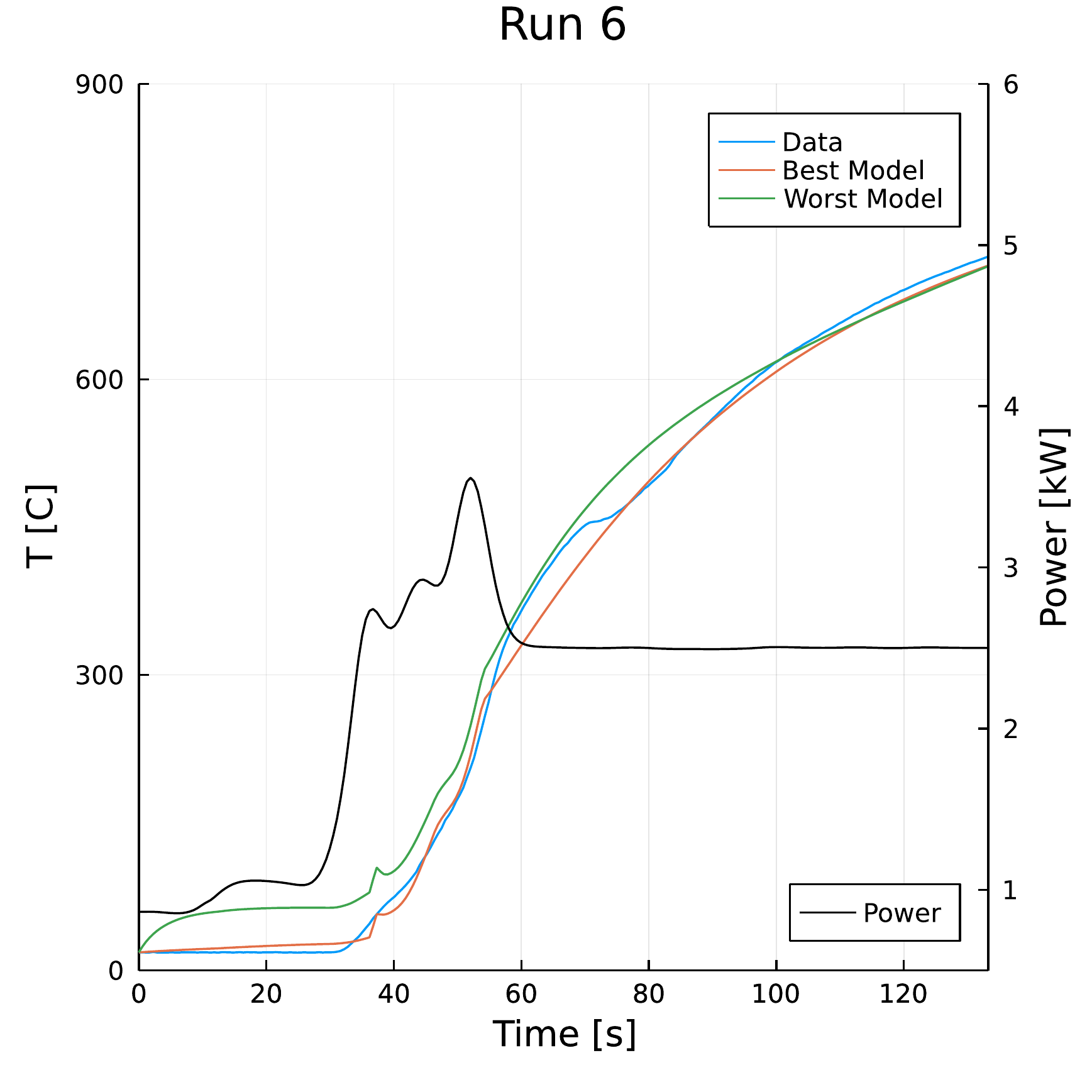}%
            \label{subfig:2b}%
        }\\
        \centering
        \subfloat[]{%
            \includegraphics[width=.48\linewidth]{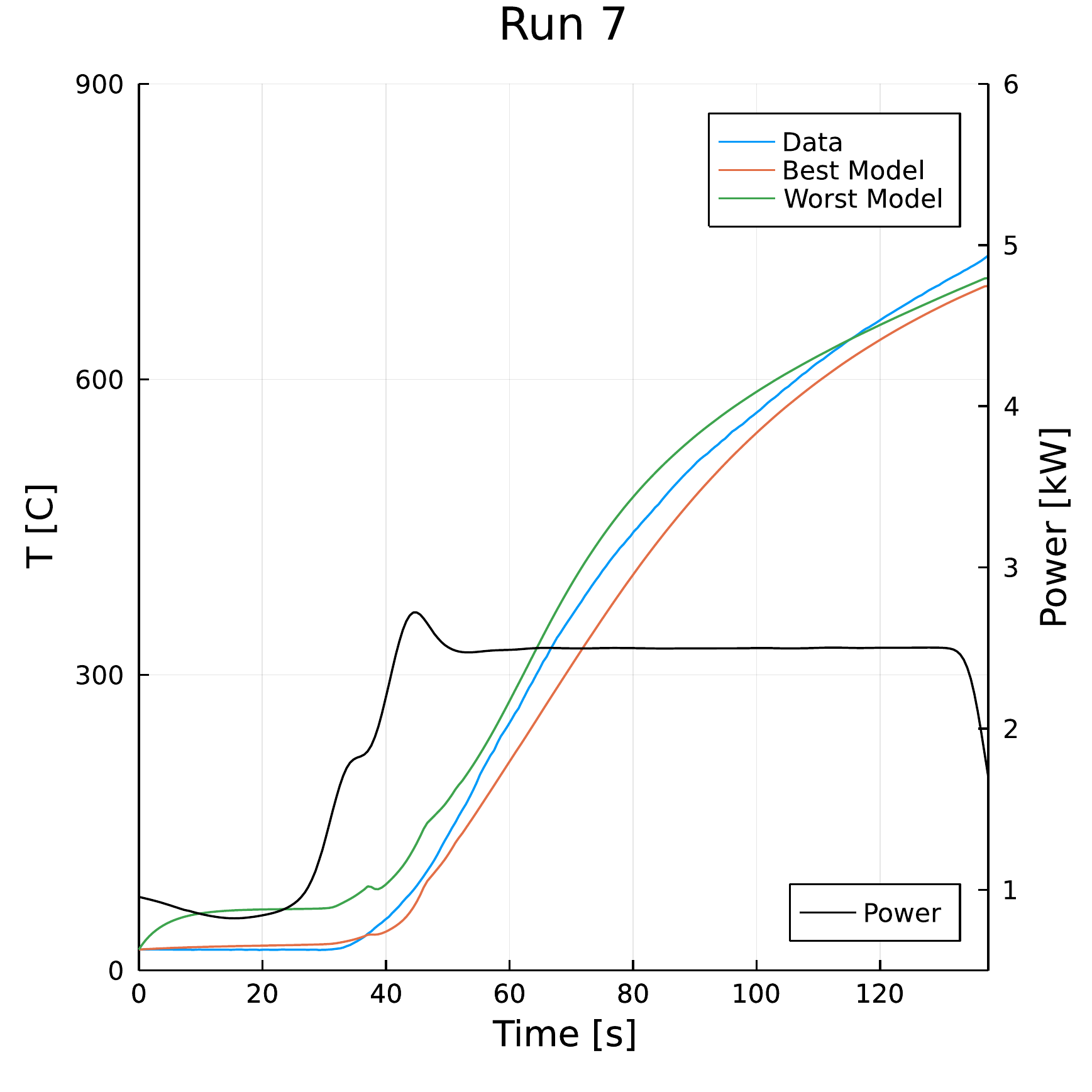}%
            \label{subfig:2c}%
        }

        \caption{Full results for experiments \#5 through \#7.}
        \label{fig:results_summary_2}
    \end{figure*}

%% If you have bibdatabase file and want bibtex to generate the
%% bibitems, please use
%%

%% else use the following coding to input the bibitems directly in the
%% TeX file.

\end{document}